\documentclass[a4paper]{article}
\usepackage{INTERSPEECH_v2}
\usepackage[utf8]{inputenc}
\usepackage{tablefootnote}
\usepackage{hyperref}
\usepackage{multirow}
\usepackage[caption=false]{subfig}

\title{Dynamic Layer Normalization\\for Adaptive Neural Acoustic Modeling in Speech Recognition}
\name{Taesup Kim$^1$, Inchul Song$^2$, Yoshua Bengio$^1$}
\address{
  $^1$MILA, Université de Montréal, Canada\\
  $^2$Samsung Advanced Institute of Technology, Republic of Korea}
\email{taesup.kim@umontreal.ca, inchul2.song@samsung.com, yoshua.bengio@umontreal.ca}

\begin{document}
\maketitle
\begin{abstract}
Layer normalization is a recently introduced technique for normalizing the activities of neurons in deep neural networks to improve the training speed and stability. In this paper, we introduce a new layer normalization technique called \textit{Dynamic Layer Normalization}~(DLN) for adaptive neural acoustic modeling in speech recognition. By dynamically generating the scaling and shifting parameters in layer normalization, DLN adapts neural acoustic models to the acoustic variability arising from various factors such as speakers, channel noises, and environments. Unlike other adaptive acoustic models, our proposed approach does not require additional adaptation data or speaker information such as i-vectors. Moreover, the model size is fixed as it dynamically generates adaptation parameters. We apply our proposed DLN to deep bidirectional LSTM acoustic models and evaluate them on two benchmark datasets for large vocabulary ASR experiments: WSJ and TED-LIUM release 2. The experimental results show that our DLN improves neural acoustic models in terms of transcription accuracy by dynamically adapting to various speakers and environments. 
\end{abstract}

\vspace{-5pt}
\section{Introduction}
Neural acoustic models have improved the transcription accuracy in speech recognition significantly over the past several years ~\cite{hinton2012deep}. Recurrent neural networks, which have cyclic connections to hold long-term temporal contextual information, are a powerful tool for modeling sequence data such as speech. In particular, the Long Short-Term Memory~(LSTM) architecture ~\cite{lstm}, which overcomes some modeling weaknesses of RNNs, has been shown to outperform DNNs and conventional RNNs for large vocabulary speech recognition \cite{bidir_lstm,sak2014long}. Despite this, neural acoustic models still suffer from the mismatch between training and testing environments. When a trained model is tested against unseen speakers or environments, its recognition accuracy can degrade substantially.

Adaptive acoustic modeling aims to adapt acoustic models to the acoustic variability across different speakers or environments. Approaches to the adaptation of neural acoustic models fall into two groups. In auxiliary feature-based adaptation \cite{saon2013speaker,gupta2014vector,tan2016speaker}, acoustic feature vectors are augmented by speaker-specific features such as i-vectors~\cite{ivector} computed for each speaker. On the other hand, in model-based adaptation ~\cite{liao2013speaker,swietojanski2014learning,swietojanski2016learning}, the model parameters are directly updated based on adaptation data. As shown in~\cite{miao2015speaker}, model-based adaptation typically brings more improvement than auxiliary feature-based adaptation. However, model-based adaptation has some drawbacks that limit its applicability in practice. For example, adaptation data needs to be gathered for each new speaker and the model size grows as the number of speakers or environments increases.

Layer normalization~\cite{layer_norm} is a recently introduced normalization method to improve the training speed and stability for various neural network models. It fixes the mean and variance of the summed inputs within each layer and a pair of trainable scaling and shifting parameters are used to adjust the normalized values. In neural style transfer~\cite{art_style_0, art_style_1, art_style_2}, a style transfer neural network is used to transfer an input image in the style of another one. Recently, it has been observed that, instead of training separate style transfer networks for each style being modeled, it is sufficient to specialize only the scaling and shifting parameters in instance normalization, which is similar to layer normalization, for each specific style~\cite{style_ln}. Motivated by this work, we investigate the use of layer normalization as a way to adapt neural acoustic models to different acoustic styles arising from different speakers and environments.    

In this paper, we introduce a new layer normalization technique called \textit{Dynamic Layer Normalization}~(DLN) for adaptive neural acoustic modeling in speech recognition. By dynamically generating the scaling and shifting parameters in layer normalization based on the input sequence, DLN adapts acoustic models to different speakers and environments. A feed forward neural network is introduced to extract from each input sequence an utterance summarization feature vector that is used to generate parameters in DLN. The whole network is jointly trained with gradient descent. Unlike other approaches in adaptive acoustic modeling, our proposed method does not require additional adaptation data or speaker information such as i-vectors. Moreover, the model does not need to be updated for each new speaker or environment as it dynamically generates adaptation parameters. We evaluate our proposed DLN applied on training deep bidirectional LSTM acoustic models on two benchmark datasets for large vocabulary ASR experiments: the Wall Street Journal~\cite{paul1992design} and TED-LIUM release 2~\cite{rousseau2014enhancing}.

The rest of this paper is organized as follows. In Section \ref{sec:related_work}, we discuss past research related to this work. In Sections \ref{sec:proposed} and \ref{sec:exp}, we describe our proposed method and experimental results, respectively. Finally conclusions follow in Section \ref{sec:conclusion}.

\vspace{-5pt}
\section{Related Work}
\label{sec:related_work}
\paragraph*{Adaptive Acoustic Modeling}
Adaptive acoustic modeling can be broadly categorized into two groups: 1)~auxiliary feature-based and 2)~model-based adaptation. Most of auxiliary feature-based adaptation methods use i-vectors~\cite{ivector} as auxiliary features in addition to input acoustic features. I-vectors can be considered as basis vectors spanning a subspace of speaker variability. In \cite{saon2013speaker,gupta2014vector}, i-vectors were used to augment the input acoustic features in DNN-based acoustic models and it was shown that appending i-vectors for each speaker resulted in improvements in the transcription accuracy. Tan et al \cite{tan2016speaker} studied the speaker-aware training of LSTM acoustic models based on i-vectors.  

On the other hand, model-based adaptation directly updates neural acoustic model parameters based on adaptation data. Liao~\cite{liao2013speaker} investigated speaker adaptation of DNN-based acoustic models using adaptation data through supervised and unsupervised adaptation and showed how L2 regularization on the speaker independent model improved generalization. In~\cite{swietojanski2014learning, swietojanski2016learning}, a speaker independent model was adapted to a specific speaker with speaker dependent parameters at each hidden layer. These parameters were estimated with adaptation data for each speaker and used to scale the hidden activations in the speaker independent model. Model-based adaptation typically brings more improvement than auxiliary feature-based adaptation as shown in \cite{miao2015speaker}. However, adaptation data needs to be collected for each new speaker and speaker-specific parameters must be maintained and estimated for each speaker, which results in an increased model size. 

\vspace{-5pt}
\paragraph*{Layer Normalization}
Layer normalization~\cite{layer_norm} was proposed to normalize the activities of neurons $x\in\mathbb{R}^{N}$ to reduce the covariate shift problem by fixing the mean and variance of $x$ within each layer in deep neural networks. 
It can be defined as a linear mapping function $LN$ with two sets of trainable parameters, scaling $\alpha$ and shifting $\beta$:
\begin{gather*}
LN\left(x;\alpha,\beta\right)=\alpha\odot\left(\frac{x-\mu}{\sigma}\right)+\beta, \nonumber \\
\quad \mu=\frac{1}{N}\sum_{i=1}^{N}{x_{i}}, \quad \quad \sigma=\sqrt{\frac{1}{N}\sum_{i=1}^{N}{\left(x_{i}-\mu_{i}\right)^{2}}} \nonumber
\end{gather*}
where $x_i$ is the $i^{th}$ element of $x$, $\mu$ and $\sigma$ are the mean and standard deviation taken across the elements of $x$, respectively. $x$ is first normalized with $\mu$ and $\sigma$ and then scaled and shifted by $\alpha\in\mathbb{R}^{N}$ and $\beta \in\mathbb{R}^{N}$. The scaling and shifting parameters are learned along with the original model parameters to restore the representation power of the network. For example, by setting $\alpha=\sigma$ and $\beta=\mu$, the original activations can be recovered. Contrary to other normalization techniques such as batch normalization~\cite{batch_norm}, it can be easily applied to recurrent neural networks since it performs exactly the same computation at training and test times. It has been shown that layer normalization is very effective at stabilizing the hidden state dynamics in recurrent neural networks.

\vspace{-5pt}
\paragraph*{Hypernetworks}
Hypernetworks~\cite{ha2016hypernetworks} were proposed to dynamically generate the weights of neural networks through weight-generating sub-networks. The whole network is trained jointly with gradient descent. When applied to recurrent neural networks, the network weights can vary across different time steps. Hypernetworks are closely related to our work in that that they also generate some of model parameters. However, the goal of hypernetworks is to relax the weight-sharing property of recurrent neural networks to control the trade off between the number of model parameters and model expressiveness.

\begin{figure}[t]
\centering
\begin{center}
\includegraphics[width=\linewidth]{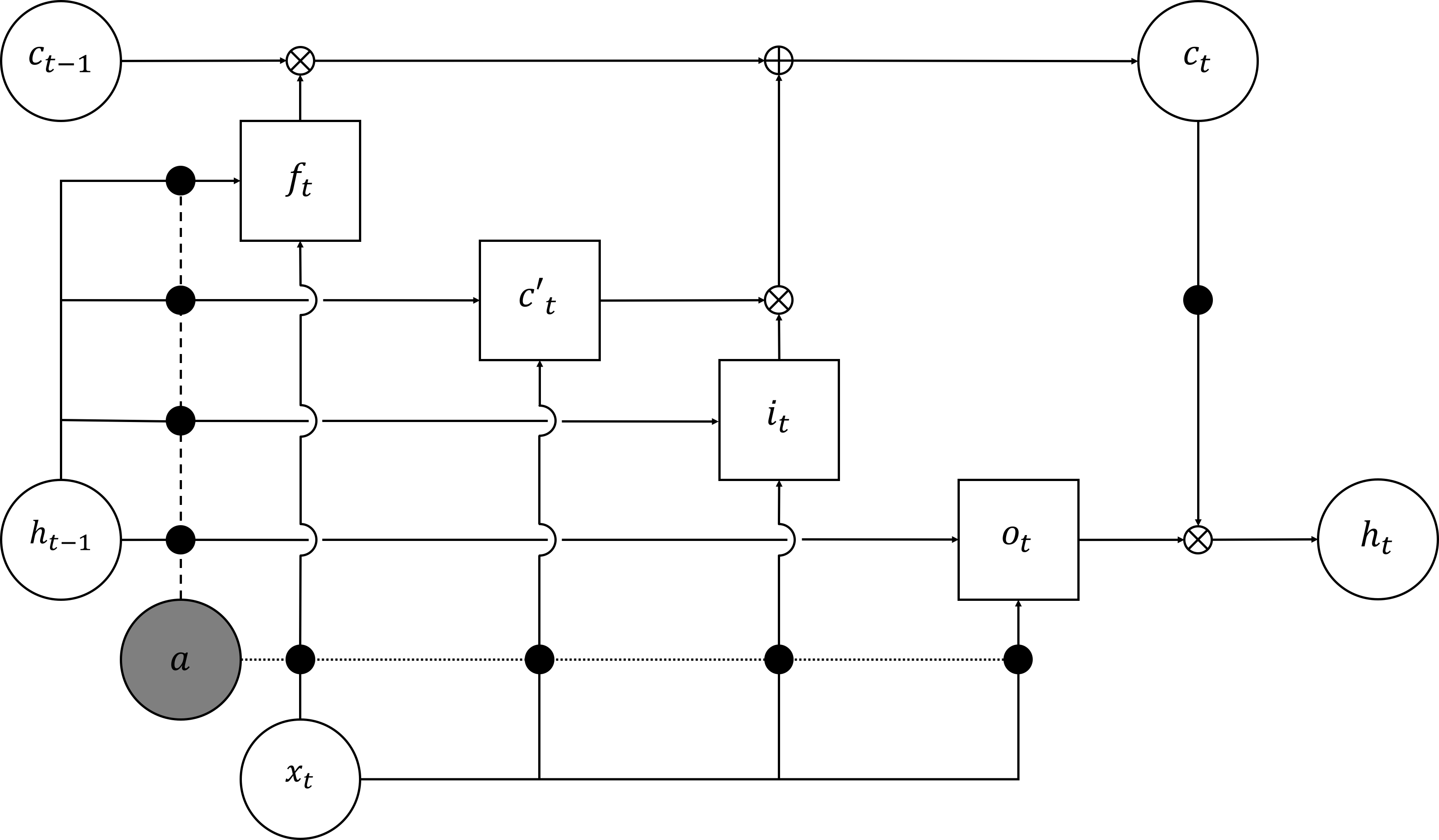}
\caption{Adaptation parameter generation process in dynamic layer normalization. The shaded circle $a$ represents the utterance summarization feature vector. As indicated by dashed lines and small black circles, the vector $a$ is used to generate the scaling and shifting parameters in layer normalization for the gates and cell state to dynamically adapt the model.}
\label{fig:dln}
\end{center}
\vspace*{-15pt}
\end{figure}

\vspace{-5pt}
\section{Proposed Model}
\label{sec:proposed}
\subsection{Baseline Architecture}
In this paper, we propose a new layer normalization technique called Dynamic Layer Normalization (DLN) and apply it to neural acoustic models based on Long Short-Term Memory (LSTM)~\cite{lstm}. LSTM has been shown to handle complex temporal dynamics in acoustic signals well. Among a number of variants of LSTM, we use the one proposed in \cite{sak2014long} called LSTM with Recurrent Project Layer (LSTMP) that has an additional recurrent projection layer to reduce the model size by mapping the hidden state into a lower-dimensional space. To stabilize the hidden state dynamics and encourage faster convergence during training, layer normalization is applied. Thus, our baseline acoustic model is defined as a composite function as follows: 
\begin{align}
i_{t}&=\sigma{\left(LN\left(W_{i}x_{t};\alpha_{i}, \beta_{i}\right)+LN\left(U_{i}h_{t-1};\alpha'_{i}, \beta'_{i}\right)\right)} \nonumber \\
f_{t}&=\sigma{\left(LN\left(W_{f}x_{t};\alpha_{f}, \beta_{f}\right)+LN\left(U_{f}h_{t-1};\alpha'_{f}, \beta'_{f}\right)\right)} \nonumber \\
o_{t}&=\sigma{\left(LN\left(W_{o}x_{t};\alpha_{o}, \beta_{o}\right)+LN\left(U_{o}h_{t-1};\alpha'_{o}, \beta'_{o}\right)\right)} \nonumber \\
c'_{t}&=\tanh{\left(LN\left(W_{c'}x_{t};\alpha_{c'}, \beta_{c'}\right)+LN\left(U_{c'}h_{t-1};\alpha'_{c'}, \beta'_{c'}\right)\right)} \nonumber \\
c_{t}&=f_{t}\odot c_{t-1} + i_{t}\odot c'_{t} \nonumber \\
h_{t}&=W_{p}\left(o_{t}\odot\tanh{\left(LN\left(c_{t};\alpha_{c}, \beta_{c}\right)\right)}\right) 
\label{eq:LSTMP}
\end{align}
where $i_{t}$, $f_{t}$, $o_{t}$, and $c_{t}$ are input gate, forget gate, output gate, and cell state, respectively. Layer normalization $LN$ is applied separately on input-to-hidden and hidden-hidden connections as proposed in \cite{layer_norm}\footnote{In our implementation, we follow the approach used in \cite{layer_norm}.
(\url{https://github.com/ryankiros/layer-norm})}. $W_{p}\in \mathbb{R}^{d'\times d}$ is a linear projection that maps the hidden state into a lower $d'$-dimensional space, where $d$ is the size of the cell state $c_{t}$. In this work, we do not use peephole connections.

In speech recognition, an input sequence $x=\left(x_1, x_2, ...,x_T\right)$ of length $T$, where $x_t$ represents a frame-level acoustic feature vector, is given at once to the system. It is therefore beneficial not only to use previous context but also future context with bidirectional recurrent neural networks (BRNN)~\cite{bidir_lstm, bidir_rnn}. Combining BRNN and LSTMP in a deep architecture, the $l^{\text{th}}$ hidden layer is defined by the forward and backward LSTMPs whose outputs are concatenated and fed into the following layer:
$$
\begin{matrix} {h}^{l}_{t}=\left[ \begin{matrix} \overrightarrow{h}^{l}_{t}  \\ \overleftarrow{h}^{l}_{t}  \end{matrix} \right] & \begin{matrix} \overrightarrow{h}^{l}_{t}=\text{LSTMP}_{\overrightarrow{\theta}^{l}}\left({h}^{l-1}_{t},\overrightarrow{h}^{l}_{t-1}\right) \\ \overleftarrow{h}^{l}_{t}=\text{LSTMP}_{\overleftarrow{\theta}^{l}}\left({h}^{l-1}_{t},\overleftarrow{h}^{l}_{t+1}\right) \end{matrix} \end{matrix}
$$
where $h^{l-1}_t$ is the output of the previous hidden layer $l-1$, $\overrightarrow{h}^l_t$ ($\overleftarrow{h}^l_t$) is the hidden state in the forward (backward) LSTMP with parameter set $\overrightarrow{\theta}$ ($\overleftarrow{\theta}$), and $h^l_t$ is the output of the hidden layer $l$,  at time step $t$. 

The output layer is defined by an affine transformation followed by a softmax function:
$$
y_{t} = \text{softmax}{\left(W_{y}h^{L}_{t} + b_{y}\right)}
$$
where $L$ is the total number of hidden layers. The output vector $y_{t}$ represents the probability distribution over all possible labels. 
In this paper, we follow the the standard approach used in hybrid
systems \cite{graves2013hybrid}. Frame-level state targets are provided on the
training set by a forced alignment given by a GMM-HMM
system. The softmax output layer has as many units as the total number of possible HMM states. The network is then trained by minimizing the negative log-likelihood of the frame-level target labels. As in \cite{graves2013hybrid}, the posterior probabilities
returned by the network are not divided by state priors during decoding.

\begin{figure}[t]
\centering
\begin{center}
\includegraphics[width=0.9\linewidth]{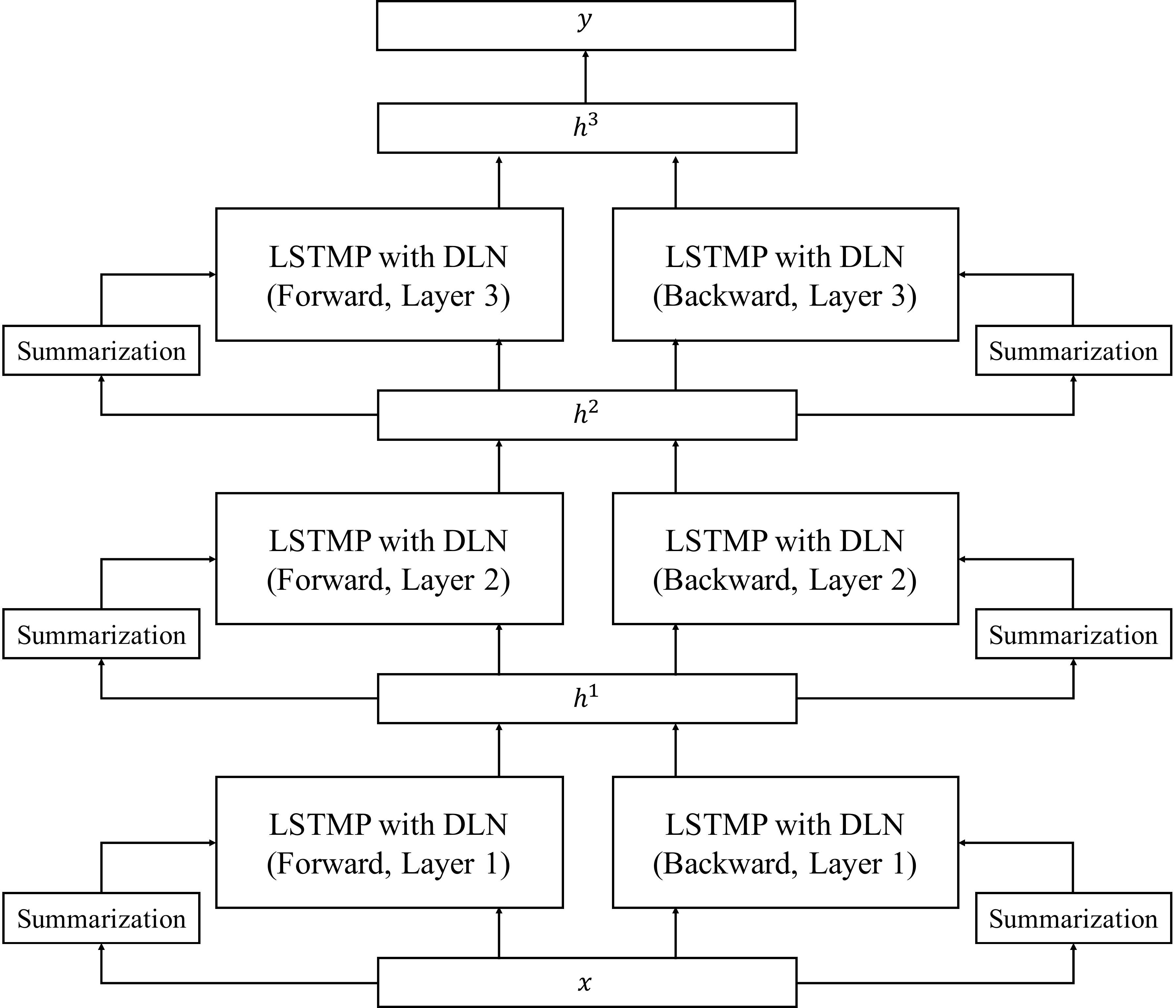}
\caption{The architecture of our proposed DLN applied to deep bidirectional LSTMP. Utterance summarization feature vectors are extracted in each layer and  direction and used to dynamically generate the parameters of layer normalization for that layer.}
\label{fig:model}
\end{center}
\vspace*{-20pt}
\end{figure}

\vspace{-5pt}
\subsection{Dynamic Layer Normalization}
Based on the deep bidirectional LSTMP with layer normalization, we propose adaptive neural acoustic models that can adapt the model based on the input sequence to handle the acoustic variability in acoustic signals due to different speakers, channels and environments. Model adaptation is done by dynamically generating the scaling and shifting parameters in layer normalization based on the input sequence rather than learning them as other parameters in neural networks. For each input sequence, different layer normalization parameters are generated, which results in effectively adapting neural acoustic models to different input sequences.

To capture the acoustic variability in different layers and directions, each forward and backward LSTMP layer has a separate utterance-level feature extractor network, which is trained jointly with the main acoustic model. The utterance summarization feature vector $a^{l}$ at layer $l$ is extracted as follows:
$$
a^{l}=\frac{1}{T}\sum_{t=1}^{T}{\tanh{\left(W^{l}_{a}h^{l-1}_{t}+b^{l}_{a}\right)}}  \quad\text{where}\quad W^{l}_{a}\in\mathbb{R}^{{p'}\times{d}}
$$
where a nonlinear transformation is applied to the output $h^{l-1}_t$ of the previous hidden layer at each time step $t$, which is followed by average pooling to obtain a fixed-length vector. We set $p'$ to be less than $d$ to reduce the computational cost of parameter generation later. The utterance summarization feature vector $a^{l}$ is then used to generate the scaling and shifting parameters, $\alpha^l_g$ and $\beta^l_g$, at layer $l$ as follows:
$$
\alpha^{l}_{g}=W^{l}_{\alpha{g}}a^l + b^{l}_{\alpha{g}} \qquad
\beta^{l}_{g}=W^{l}_{\beta{g}}a^l + b^{l}_{\beta{g}}
$$
where $g$ is one of $\lbrace i, f, o, c'\rbrace$ in Equation~\ref{eq:LSTMP}. The process of dynamically generating adaptation parameters based on utterance summarization feature vectors is depicted in Figure~\ref{fig:dln}.

Figure~\ref{fig:model} shows the architecture of a deep bidirectional LSTMP with DLN. Note that DLN does not need any additional adaptation data or speaker information such as i-vectors. Moreover, the model size does not change because adaptation parameters are dynamically generated.

In order to extract more discriminative utterance summarization features that represent various factors in acoustic signals, we add a penalty term, $\text{L}_{\text{var}}$, to the loss to encourage each feature $a^l_i$ in the utterance summarization feature vector $a^l$ to be highly varied across the utterances within each mini-batch during training:
\begin{equation}\label{eq:reg}
\text{L}_{\text{var}}=-\lambda \frac{1}{L}\sum_{l=1}^{L}{\frac{1}{p'}\sum_{i=1}^{p'}{\text{var}(a^l_{i})}}
\vspace*{-5pt}
\end{equation}
where the variance $\text{var}(\cdot)$ is computed over the minibatch, $L$ is the total number of hidden layers, and $\lambda$ is a hyperparameter that weights the contribution of the penalty relative to the loss.

\vspace{-5pt}
\section{Experiments}
\label{sec:exp}
\subsection{Datasets}
We evaluate our proposed methods on two benchmark datasets for large vocabulary automatic speech recognition experiments: the Wall Street Journal~(WSJ) corpus~\cite{paul1992design} and TED-LIUM corpus release 2 ~\cite{rousseau2014enhancing}. The WSJ corpus primarily consists of read speech with texts drawn from a machine-readable corpus of Wall Street Journal news text. The TED-LIUM corpus release 2 is composed with segments of public talks extracted from the TED website. The collective summary of statistics for each corpora is given in Table~\ref{tab:corpus}.  

For both datasets, each frame in the acoustic signal is represented by 40 log Mel-filterbank outputs (plus energy), together with their first and second derivatives. Each utterance is then represented as a sequence of frames where the size of each frame is 123.
\begin{table}[t]
  \caption{Corpus Statistics}
  \label{tab:corpus}
  \centering
  \resizebox{0.465\textwidth}{!}{
  \begin{tabular}{l l l l l}
    \toprule
    \textbf{Corpus} &  & \textbf{Train} & \textbf{Dev} & \textbf{Test}\\
    \midrule
    \multirow{2}{*}{WSJ} & \# Utterances & 37416 (81h) & 503 & 333 \\ 
                         & \# Speakers & 283 & 10 & 8 \\
    \midrule
    \multirow{2}{*}{\begin{tabular}[c]{@{}l@{}}TED-LIUM\\Release 2\end{tabular}} & \# Utterances & 92973 (212h) & 507 & 1155 \\
							  & \# Speakers & 5076 (1495) & 38 (8) & 59 (11) \\    
    \bottomrule
  \end{tabular}}
\vspace*{-10pt}
\end{table}
\vspace{-5pt}
\subsection{Network Architecture and Training}
All neural acoustic models in the experiments have three bidirectional LSTMP hidden layers, with 512 LSTM cells and 256 recurrent projection units in each of the forward and backward directions. Layer normalization is applied to all layers as in Equation~\ref{eq:LSTMP}, and  only one bias term $\beta$ (shifting parameter) for each $g\in\lbrace i_t, f_t, o_t, c'_t \rbrace$ is used in our implementation to avoid unnecessary redundancy between $\beta_{i}$ and $\beta'_{i}$. The Adam optimizer~\cite{DBLP:journals/corr/KingmaB14} is used for training models with the initial learning rate set to 0.001. The mini-batch size is set to 16. All weights are initialized by orthogonal initialization \cite{henaff2016orthogonal} and biases are set to zero. To reduce the computational cost of generating parameters in DLN, the size of utterance summarization feature vector, $p'$, is set to 64. All models are implemented in Theano~\cite{2016arXiv160502688short} using the Lasagne neural network library~\cite{lasagne}.

\begin{table}[t]
\centering
\caption{Experimental results}
\label{tab:results}
\vspace{-5pt}
\subfloat[Wall Street Journal]{
\resizebox{0.465\textwidth}{!}{
\begin{tabular}{lrrr}
\toprule
\multicolumn{1}{c}{\multirow{2}{*}{\textbf{Model}}} & \multicolumn{1}{c}{\multirow{2}{*}{\textbf{Size}}} & \multicolumn{1}{c}{\textbf{Dev FER}} & \multicolumn{1}{c}{\textbf{Test FER}} \\
\multicolumn{1}{c}{} & \multicolumn{1}{c}{} & \multicolumn{1}{c}{\textbf{Dev WER}} & \multicolumn{1}{c}{\textbf{Test WER}} \\
\midrule
\multirow{2}{*}{LSTMP w/ LN} & \multirow{2}{*}{10.44M} & 22.68\% & 23.71\% \\
 &  & 7.26\% & 4.50\% \\
\multirow{2}{*}{LSTMP w/ DLN} & \multirow{2}{*}{12.94M} & 21.81\% & 23.35\% \\
 &  & 7.09\% & 4.63\% \\
 \bottomrule
\end{tabular}}}
\vspace*{-5pt}
\\
\subfloat[TED-LIUM Release 2]{
\resizebox{0.465\textwidth}{!}{
\begin{tabular}{lrrr}
\toprule
\multicolumn{1}{c}{\multirow{2}{*}{\textbf{Model}}} & \multicolumn{1}{c}{\multirow{2}{*}{\textbf{Size}}} & \multicolumn{1}{c}{\textbf{Dev FER}} & \multicolumn{1}{c}{\textbf{Test FER}} \\
\multicolumn{1}{c}{} & \multicolumn{1}{c}{} & \multicolumn{1}{c}{\textbf{Dev WER}} & \multicolumn{1}{c}{\textbf{Test WER}} \\
\midrule
\multirow{2}{*}{LSTMP w/ LN} & \multirow{2}{*}{10.81M} & 24.05\% & 24.68\% \\
 &  & 14.18\% & 13.50\% \\
\multirow{2}{*}{LSTMP w/ DLN} & \multirow{2}{*}{13.32M} & 23.27\% & 23.82\% \\
 &  & 13.62\% & 12.82\% \\
 \bottomrule
\end{tabular}}}
\vspace*{-25pt}
\end{table}

\vspace{-5pt}
\subsection{Results and Discussion}
\paragraph*{Wall Street Journal Experiments}
We follow the standard Kaldi recipe s5 \cite{povey2011kaldi} for preparing speech data. 
A baseline GMM-HMM system is trained on the 81 hours training set (train-si284) by Kaldi recipe tri4b, which consists of LDA preprocessing of data, with MLLT and SAT for adaptation. We then generate a forced alignment to obtain frame-level targets. There are 3436 triphone states in total. We use the dataset test-dev93 as the development set and test-eval92 as the test set.

Table~\ref{tab:results} (a) shows the results of the experiments reported in terms of Frame Error Rates (FER) and Word Error Rates (WER). As shown in the table, the model trained with DLN outperforms the baseline model for both of the dev and test sets in terms of FER, but the improvement was not similarly shown on the WER. This is suspected that the WSJ corpus has a small number of speakers and was recorded under clean conditions that other environmental factors wouldn’t effect the acoustic variability. Moreover, the proposed regularizer for DLN does not help much, so $\lambda$ is set to 0.

\vspace{-5pt}
\paragraph*{TED-LIUM Experiments}
Speech data is prepared by following the standard Kaldi recipe s5\_r2. The speakers are split up into 3-minute chunks for better generalization and fast per-utterance decoding.
Table~\ref{tab:corpus} shows both the increased number of speakers and the original number of speakers in parentheses.
We first train a baseline GMM-HMM system on the 212 hours training set by Kaldi recipe tri3 and generate a forced alignment with 4174 triphone states in total. For DLN, we set $\lambda$ to 10, which gave the best result on the dev set.

The experimental results are shown in Table~\ref{tab:results} (b). The larger TED-LIUM corpus contains far more utterances and speakers than the WSJ corpus and was recorded from various environments. As shown in the table, the model trained with DLN is able to adapt to the high variability in the corpus and outperforms the baseline model on both of the dev and test sets.

\begin{figure}[t]
\centering
\begin{center}
\subfloat[First layer]{\includegraphics[width=0.5\linewidth]{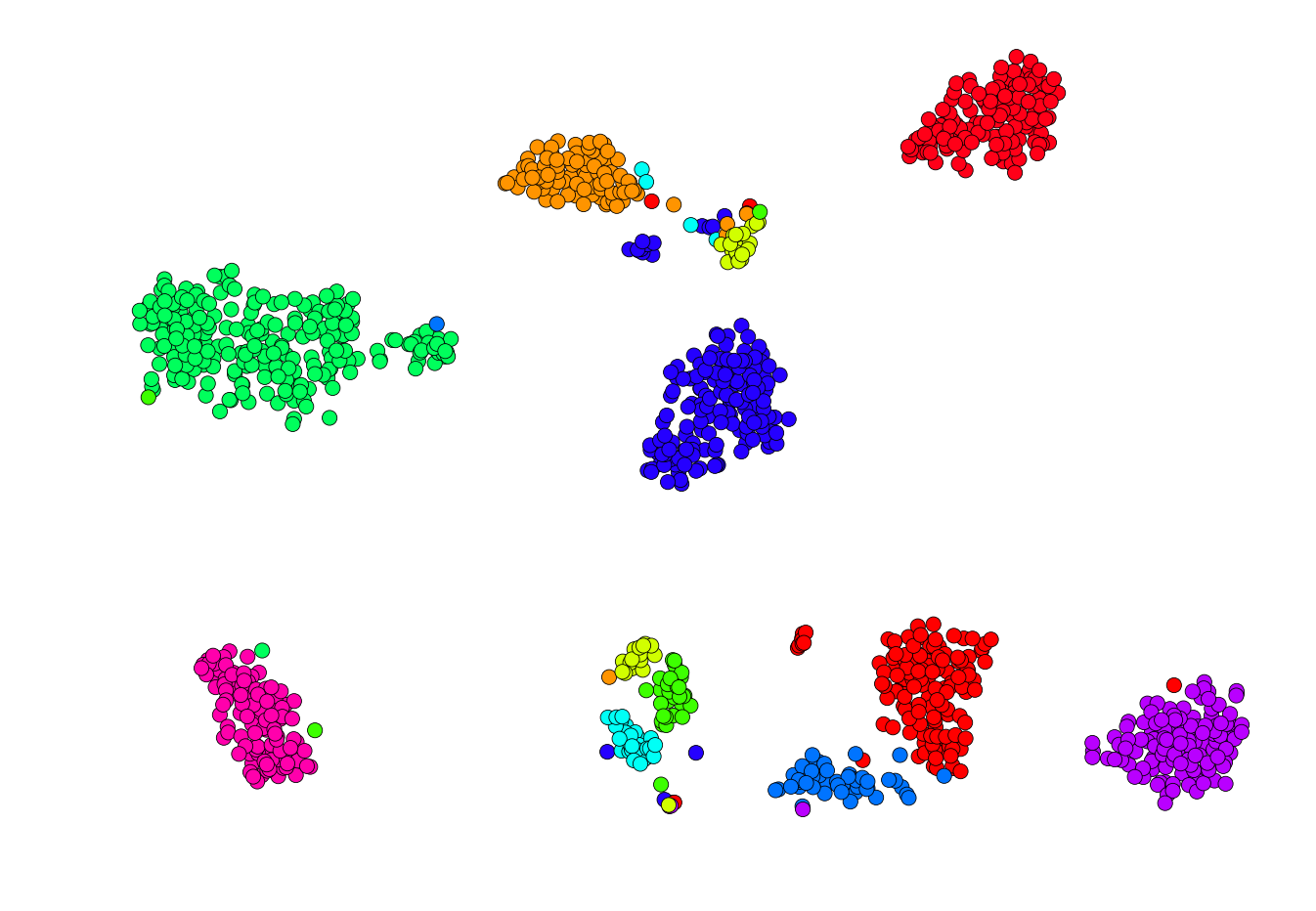}}
\subfloat[Third layer]{\includegraphics[width=0.5\linewidth]{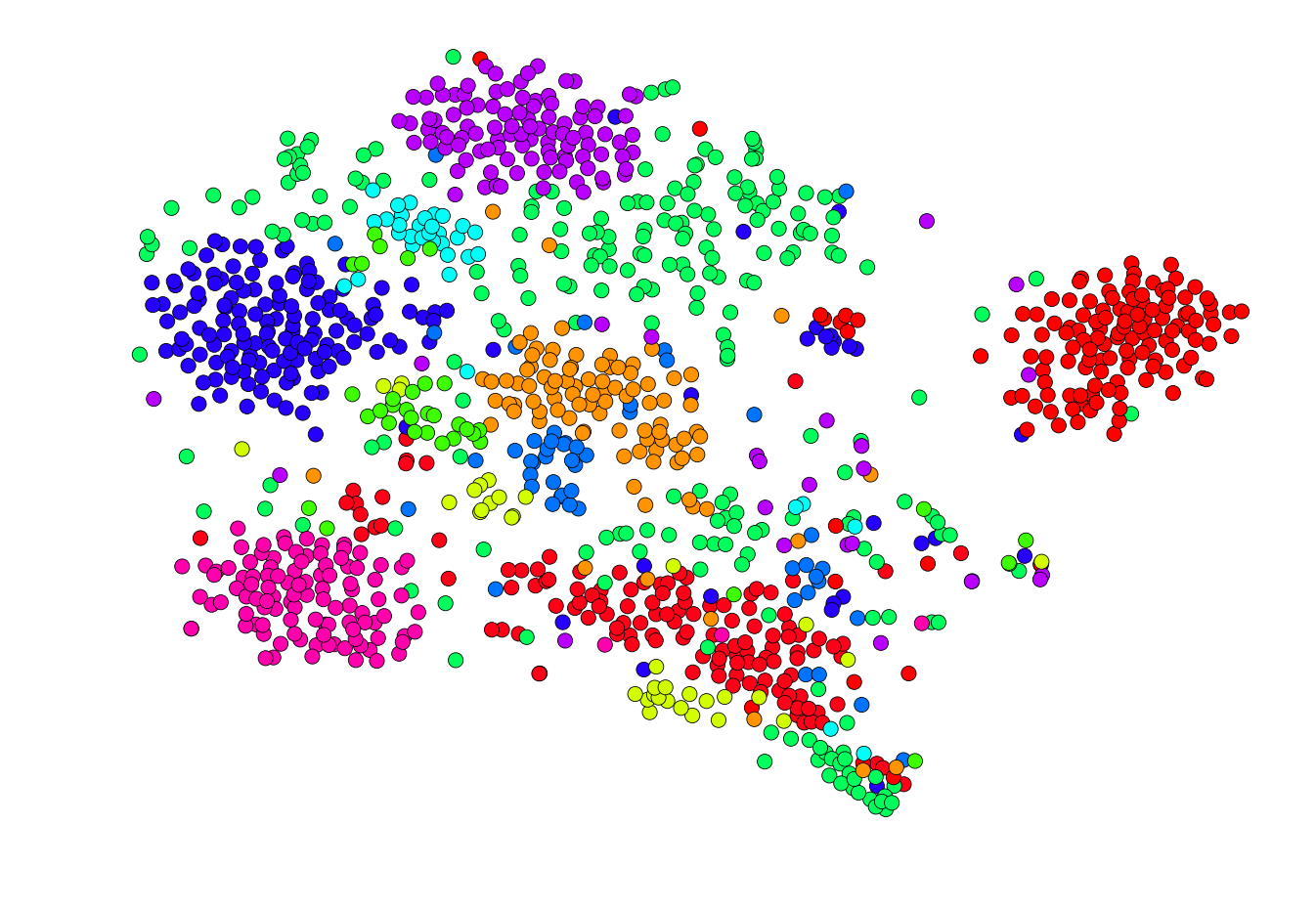}}
\caption{The utterance summarization feature vectors extracted from (a) the first layer and (b) the third layer in our proposed model are plotted using \textit{t-SNE} in 2D-space. Each color represents a distinct speaker identity. The test set of the TED-LIUM corpus contains 11 different speakers.}
\label{fig:feat}
\end{center}
\vspace*{-25pt}
\end{figure}

\vspace{-5pt}
\paragraph*{Discussion}
We visualize the utterance summarization feature vectors by using \textit{t-SNE}~\cite{ictdbid:2777}, which is a technique for visualizing high dimensional data into 2D space. We use the test set of the TED-LIUM corpus and plot the feature vectors from the first and third layers. Figure~\ref{fig:feat} shows how the utterance summarization feature vectors are clustered and correlated with the speaker identity. The feature vectors from the first layer are clustered and highly correlated to speaker identities such that the number of clusters are similar to the original number of speakers, which is 11, even though no speaker information is used. On the other hand, the feature vectors from the third layer are more scattered. This can be interpreted that the feature vectors from the lower layers, which are closer to the input acoustic signals, represent speaker-related features and those from the higher layers are related to other factors.

For both of the datasets, we trained larger baseline models by adding one more layer to make their sizes similar to those trained with DLN. However, their performances degraded due to overfitting. On the other hand, DLN utilizes an increased model capacity effectively for generating adaptation parameters that led to performance improvements and empirically showed faster convergence as well during training.

\vspace{-5pt}
\section{Conclusions}
\label{sec:conclusion}
In this paper, we have proposed a new layer normalization technique called DLN for adaptive neural acoustic model training in speech recognition. By dynamically generating scaling and shifting parameters for layer normalization based on the input sequence, DLN adapts neural acoustic models to various speakers and environments. Unlike other adaptive acoustic models, DLN does not require additional adaptation data or contextual information such as speaker identity. In addition, the model size does not increase as it dynamically generates adaptation parameters. We have shown through experimental evaluation that DLN improves neural acoustic models in terms of transcription accuracy.

As future work, we plan to investigate other ways to extract more useful summarization features from the input sequence to help generate adaptation parameters. 

\vspace{-5pt}
\section{Acknowledgements}
The authors would like to thank Aaron Courville for valuable comments and also acknowledge the support of the following agencies for
research funding and computing support: NSERC, Samsung, Calcul Quebec, Compute Canada, the Canada Research Chairs and CIFAR.
\bibliographystyle{IEEEtran}

\bibliography{mybib.bib}

\end{document}